\documentclass[letterpaper, 10pt, journal]{IEEEtran}

\usepackage{graphics} 
\usepackage{epsfig} 
\usepackage{mathptmx} 
\usepackage{amsmath} 
\usepackage{bm}
\usepackage{stfloats}
\usepackage{upgreek}
\usepackage{color}
\usepackage{makecell}
\usepackage[caption=false,font=normalsize,labelfont=sf,textfont=sf]{subfig}

\begin{document}

\title{Design and trajectory tracking control of CuRobot: A Cubic Reversible Robot}

\author{Kai Yang$^{1}$, 
        Jiahui Wang$^{1}$, 
        Yuchen Weng$^{1}$, 
        Baolei Wu$^{1}$,
        Fuqiang Li$^{1}$,
        Jihong Zhu$^{2}$, \\
        Jun Wang$^{1}$, ~\IEEEmembership{Member,~IEEE}

\thanks{$^{1}$Kai Yang, Jiahui Wang, Yuchen Weng, Baolei Wu, Fuqiang Li and Jun Wang are with School of Information and Control Engineering, China University of Mining and Technology, Xuzhou 221116, China
        {\tt\small (e-mail:yk267x@163.com, wjh371522@163.com, yuchenwengwyc@gmail.com, blwu@cumt.edu.cn, lifuqiang@cumt.edu.cn, jrobot@126.com)}}
\thanks{$^{2}$Jihong Zhu is with Department of Precision Instrument, Tsinghua University, Beijing 100084, China
        {\tt\small (e-mail:zjhcumtedu@163.com)}}
}
\maketitle

\begin{abstract}

    In field environments, numerous robots necessitate manual intervention for restoration of functionality post a 
    turnover, resulting in diminished operational efficiency. This study presents an innovative design solution for 
    a reversible omnidirectional mobile robot denoted as CuRobot, featuring a cube structure, thereby facilitating 
    uninterrupted omnidirectional movement even in the event of flipping. The incorporation of eight conical wheels 
    at the cube vertices ensures consistent omnidirectional motion no matter which face of the cube contacts the 
    ground. Additionally, a kinematic model is formulated for CuRobot, accompanied by the development of a trajectory 
    tracking controller utilizing model predictive control. Through simulation experiments, the correlation between 
    trajectory tracking accuracy and the robot's motion direction is examined. Furthermore, the robot's proficiency 
    in omnidirectional mobility and sustained movement post-flipping is substantiated via both simulation and prototype 
    experiments. This design reduces the inefficiencies associated with manual intervention, thereby increasing the 
    operational robustness of robots in field environments.
\end{abstract}

\begin{IEEEkeywords}
     Mechanism Design, Kinematics, Omnidirectional motion, trajectory tracking.
\end{IEEEkeywords}
\section{INTRODUCTION}

Mobile robots are already capable of assisting humans across various domains. However, in challenging terrains, these robots may encounter issues such as tipping over, impeding their ability to function autonomously, and necessitating human intervention, which, in turn, restricts their practical applications \cite{ref1,ref2}. Thus, the capability to resume operations after tipping over in complex terrains holds substantial practical significance \cite{ref3}. Conventional wheeled robots \cite{ref4,ref5,ref6} and tracked robots \cite{ref7,ref8} lack the capacity to sustain mobility after tipping over. In response, researchers have introduced a spectrum of diverse robot configurations as potential solutions, which can be broadly categorized into three main groups: 1) spherical robots \cite{ref9,ref10,ref11,ref12,ref13,ref14}, 2) RHex robots and their improved versions \cite{ref15,ref16}, and 3) legged robots \cite{ref17,ref18,ref19,ref20,ref21}.

Spherical robots exhibit remarkable adaptability to significant environmental changes, alleviating concerns about potential disruptions in movement due to tipping over. However, a notable drawback of spherical robots lies in the intricate nature of their mechanical structures. This complexity can render the manufacturing and maintenance of spherical robots more challenging and cost-intensive. Conversely, the precise control of spherical robots for accurate position and orientation tracking can be a demanding task, owing to their distinct mobility capabilities and the requirement for precise sensor data.

RHex is an autonomous robot design based on a hexapod with compliant legs and one actuator per leg. It is designed for mobility on rough terrain and can operate right-side-up or upside down and one of the versions has shown good mobility over a wide range of terrain types at speeds exceeding five body lengths per second (2.7 m/s), climbed slopes exceeding 45 degrees, swims, and climbs stairs. RHex robots have a limitation in their inability to perform lateral movement.

Legged robots can re-establish their upright position following a fall by orchestrating the movement of their limbs. However, this necessitates the development of distinct recovery strategies tailored to various falling postures \cite{ref22,ref23}, demanding a controller with high computational efficiency.

To overcome these limitations, we have designed a robot named CuRobot which can continue to move even after tipping over. The robot has an overall cubic structure and has three degrees of freedom for omnidirectional motion regardless of which side it lands on. Building upon the structural design, we conducted a comprehensive analysis of its kinematic model and designed the trajectory tracking controller using model predictive control \cite{ref31,ref32} (MPC). Ultimately, we validated the distinctiveness of 
its motion through a series of experiments.

A brief description of CuRobot's structure and a physical prototype is given in Section \uppercase\expandafter{\romannumeral2}.
Section \uppercase\expandafter{\romannumeral3} establishes the kinematic model and designs the controller.
Experimental results are analyzed and discussed in Section \uppercase\expandafter{\romannumeral4}.
Concluding remarks are given in Section \uppercase\expandafter{\romannumeral5}.

\section{ROBOT DESIGN}
\begin{figure*}[ht]
        \centering
        \includegraphics[scale=0.4]{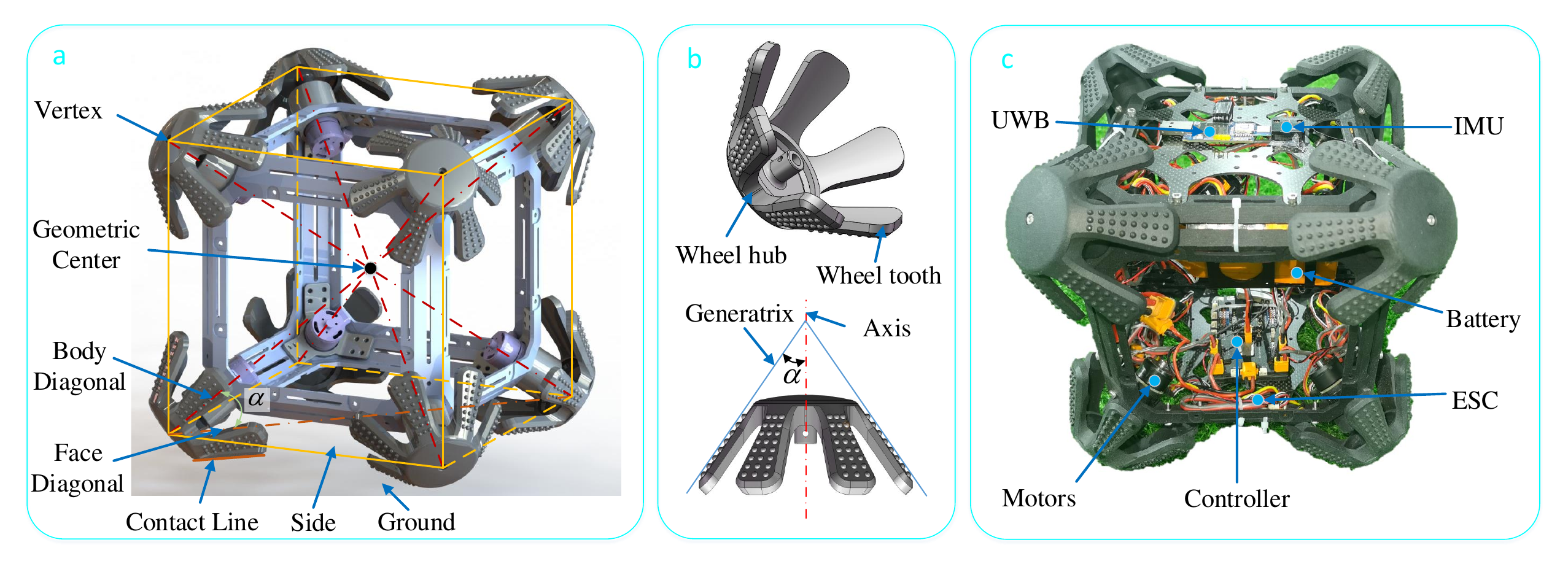}
        \caption{Overview of CuRobot. (a) Structure design of CuRobot; (b) Design of the cone wheel; (c) Prototype of CuRobot.}
        \label{fig:robot_design}
\end{figure*}

We employ a hexahedral, i.e., cube, structure to facilitate continuous motion even after flipping. In this scenario, the robot initially lands on one of its faces and lands on another once flipped. To ensure that the robot maintains its four-wheel drive omnidirectional capability regardless of which side makes contact with the ground, the wheels are positioned at the vertices of the cube, with their rotating axes aligned with the body diagonals of the cube, as shown in Fig. \ref{fig:robot_design}a. When the robot flips 90°, two out of the four wheels that were initially in contact with the ground lose contact, while the two wheels on the adjacent side, which were initially not in contact with the ground, come into contact. Consequently, four wheels are consistently in contact with the ground, regardless of which side of the robot is oriented toward the ground.

For the side of the CuRobot facing the ground, it is essential to establish a linear contact between the four wheels located on that side and the ground. This arrangement maximizes the contact area between the wheels and the ground. Geometrically, the contact line runs parallel to the side of the cube and exactly parallel to the face diagonal. By taking the contact line as the generatrix, rotating it around the wheel axis yields the conical surface, which forms the wheel surface. So, the conic wheel is designed with an envelope surface in the form of a circular truncated cone, a portion of the conical surface. Fig. \ref{fig:robot_design}b shows the angle, denoted as $\alpha$, between the generatrix and the circular axis. This angle is equal to the angle between the body diagonal and the side of the cube, which can be expressed as $arctan(\frac{1}{\sqrt{2}})$. Therefore, based on the wheel shape and mounting mode, there are three lines on all wheels, respectively parallel to the three mutually perpendicular sides intersecting at the locations where the wheels are positioned.

Ordinary wheels exhibit static friction with the ground during travel. However, sliding friction arises between the conic wheels and the ground when the four conical wheels cooperate to accomplish the desired motion. Generally, it is difficult to ensure definite relative sliding between the four driving wheels and the ground, which leads to the uncertainty of the bulk motion of CuRobot. To reduce the motion uncertainty caused by sliding friction force and enhance the robot's ability to overcome obstacles, the wheels are designed with a toothed structure, where the width of the tooth and backlash are uniform. The wheel consists of two parts: the hub and the tooth. One end of the tooth connects to the hub, while the other end protrudes outward, as shown in Fig. \ref{fig:robot_design}b. To prevent the tooth structure from causing bumps during robot movement, the radius of the hub is equal to the radius of the tooth's root. This ensures that the hub maintains continuous contact with the ground while the tooth has intermittent contact.

Compared to other robots that can be reversed, this robot with a novel structure has some special features
in its motion characteristics, as shown in TAB. \ref{compare}.

\begin{table}[!htb]
        \centering
        \caption{Comparison of the motion performance of different reversible robots}  
        \label{compare}
        \begin{tabular}{ccccc}
        \hline
                           & \makecell{omnidirectional\\ motion} & \makecell{obstacle\\ crossing} & invertibility  \\ \hline
 \makecell{sphere robot} & $\surd$                             &     bad                        & omnidirectional \\
        RHex               & $\times$                            &     better                     &   sagitta      \\
\makecell{legged robot}  & $\surd$                             &    best                        & omnidirectional \\
        proposed           & $\surd$                             &   good                         & omnidirectional \\
        \hline
        \end{tabular} 
\end{table}

Fig. \ref{fig:robot_design}c shows the physical prototype of CuRobot, powered by a 6S lithium battery boasting a capacity of 2,200 mAh. The robot's controller is DJI's RoboMaster Type A control board. Each of CuRobot's wheels is autonomously driven by a DC brushless gear motor, specifically the RoboMaster M2006 P36, designed to operate at a rated voltage of 24 V. The motor's operation is overseen by the RoboMaster C610 electronic speed controller, capable of achieving a maximum speed of 500 rpm and sustaining a maximum torque of 1,000 mNm. Furthermore, the robot is equipped with WIT's WT901CM attitude sensor and HaoruTech's UWB location module, which integrates a DW1000 radio frequency chip. CuRobot's body and wheels are manufactured using 3D printing, with the material being high-performance nylon.

\begin{table}[!htb]
        \centering
        \caption{WEIGHT OF EACH COMPONENT}
        \label{weight}
        \begin{tabular}{cccc}
        \hline
        component          &  weight(g)  & number  &  sum \\ \hline
        frame              &  115        &  1      &  245   \\
        battery \& bracket &  363        &  1      &  363   \\
        motors             &  90         &  8      &  720   \\
          ESC              &  17         &  8      &  136   \\
        wheel              &  48         &  8      &  384   \\
        controller         &  54         &  1      &  54   \\
        receiver           &  11         &  1      &  11   \\ 
        IMU                &  6          &  1      &  6   \\
        UWB                &  55         &  1      &  55   \\
        fastenings \& wire &             &         &  284  \\
        $\bm{total}$       &             &         &  $\bm{2258}$  \\
        \hline
        \end{tabular}
\end{table}

\section{CONTROL}
\begin{figure*}[!t]
        \centering
        \subfloat[Schematic diagram of kinematics analysis]{\includegraphics[scale=0.75]{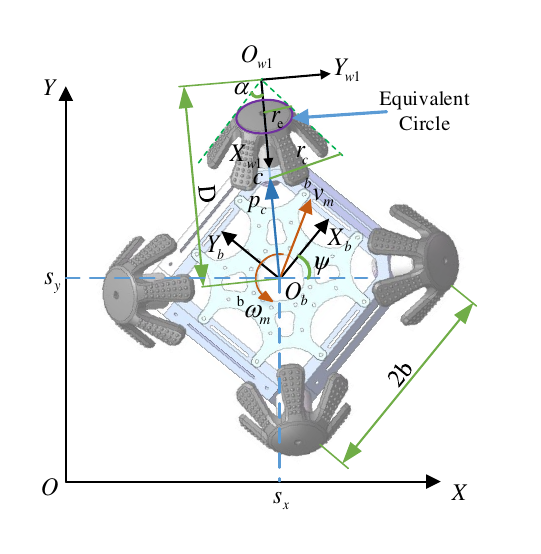}
        \label{fig:kinematic}}  
        \hfil
        \subfloat[Definition of initial body coordinate system]{\includegraphics[scale=0.42]{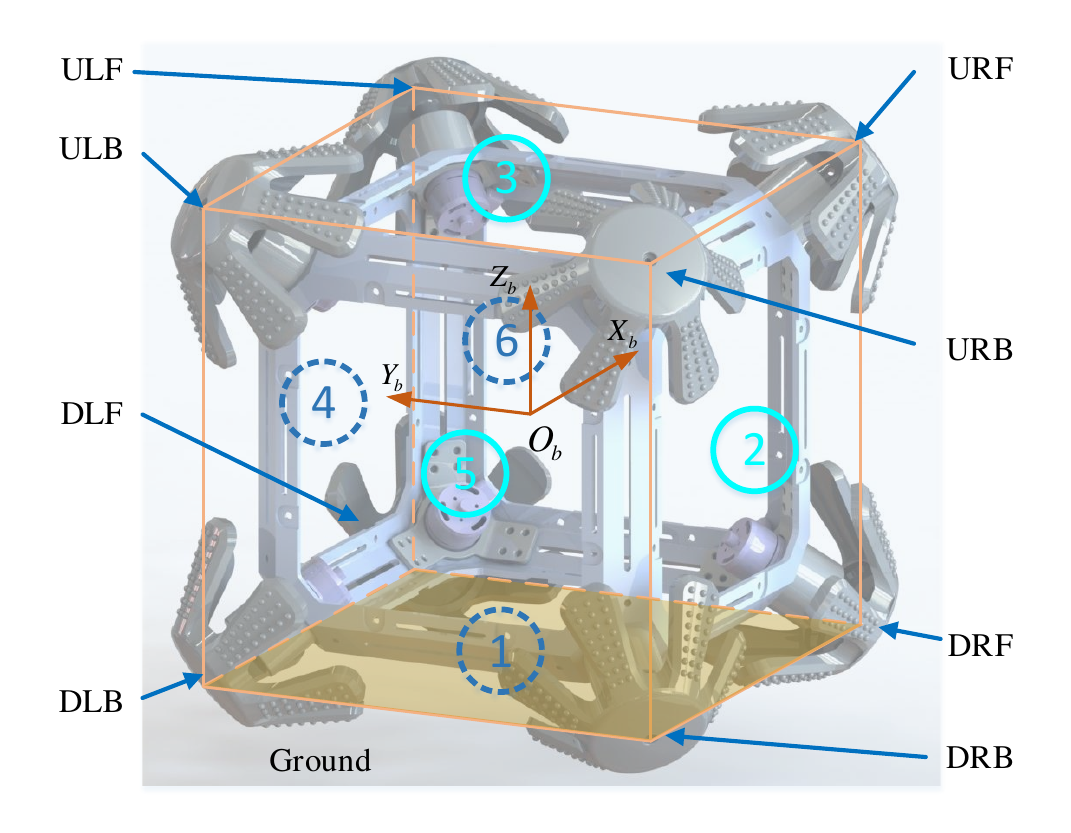}
        \label{fig:face}}
        \caption{Schematic diagram of theoretical models.}
        \label{fig:motion}
\end{figure*}

In this section, the kinematic model of CuRobot is established, and the trajectory tracking controller is designed. We set three frames for later analysis: the world frame $XOY$, the body frame $X_bO_bY_b$, and the wheel frame $X_{wi}O_{wi}Y_{wi}$, where $i=1,2,3$, and $4$, represents the serial number of the four driving wheels (i.e., wheels contacting the ground). The origin of the world frame is set on the ground and the origin of the body frame, typically placed at the center of gravity of the robot, is set at the projection of the geometric center of the robot on the ground. The origin of the $X_{wi}O_{wi}Y_{wi}$ frame is located at the virtual vertex of the circular envelope surface of the wheel. The $O_{wi}X_{wi}$ coincides with the projection of the rotation axis of the wheel on the horizontal plane, and its direction points from $O_{wi}$ to $O_b$. $O_{wi}Y_{wi}$ lies in the horizontal plane and is perpendicular to $O_{wi}X_{wi}$, as shown in Fig. \ref{fig:kinematic}. 

\subsection{Kinematic Model}
The kinematic model indicates the relationship between the motion of the robot's centre of mass and the rotational speed of the driving wheels, and coordinating the rotation of the four driving wheels allows the robot to achieve omnidirectional motion. Fig. \ref{fig:kinematic} shows the kinematic analysis model. During the course of movement, slip may occur between the conical wheels and the ground due to variations in the wheel radius, which can hinder the accurate calculation of the wheels' rotational speed. In practical applications, it is necessary to evaluate the actual rotation radius. In this paper, we assumed the circle where the hub is located as the equivalent circle, and the radius of this equivalent circle represents the wheel's equivalent radius, denoted as $r_e$. 

First, the centroid velocity of the robot in the $XOY$ frame is transformed into the $X_bO_bY_b$ frame. This can be expressed as:

\begin{equation}
        ^b\bm{\dot{\chi}} = \bm{T}^{-1}\bm{\dot{\chi}} ,
    \end{equation}
where $ \bm{\chi}={\begin{bmatrix} s_x & s_y & \psi \end{bmatrix}}^T$ is the position and yaw angle of the robot, and 
        $\bm{T}=\begin{bmatrix}   
                    cos\psi & -sin\psi & 0\\
                    sin\psi & cos\psi & 0\\
                    0 &  0 & 1
                \end{bmatrix}$             
is the transformation matrix from $X_bO_bY_b$ to $XOY$. The transformation of the reference system only encompasses rotational changes, as the representation of velocity in distinct coordinate systems is solely dependent on the relative orientation of the two coordinate systems, rather than their relative positions. The planar rotation transformation matrix is originally two-dimensional, yet it is expanded into three dimensions to accommodate the inclusion of the yaw angle $\psi$.

Next, the speed of the equivalent circle of each wheel is calculated according to the motion of the center 
of mass, and the results are expressed in $X_bO_bY_b$ frame as follows:

\begin{equation}
        ^b\bm{v}_{i,e} = \mathbf{T}_1 {^b\dot{\bm{\chi}}} + (\mathbf{T}_2 {^b\dot{\bm{\chi}}})\times \mathbf{p}_{i,e},
\end{equation}
where $\mathbf{T}_1=\begin{bmatrix} 1 & 0 & 0\\ 0 & 1 & 0\\ 0 & 0 & 0 \end{bmatrix}$, 
       $\mathbf{T}_2=\begin{bmatrix} 0 & 0 & 0\\ 0 & 0 & 0\\ 0 & 0 & 1 \end{bmatrix}$,
and $\mathbf{p}_{i,e}$ represents the vector pointing from $O_b$ to the contact point $e$ between the equivalent 
circle and the ground, and
$\mathbf{p}_{1,e}={\begin{bmatrix}\text{b} & \text{b}&0 \end{bmatrix}}^T$, $\mathbf{p}_{2,e}={\begin{bmatrix}\text{b} & -\text{b}&0 \end{bmatrix}}^T$, $\mathbf{p}_{3,e}={\begin{bmatrix}-\text{b} & -\text{b}&0 \end{bmatrix}}^T$, $\mathbf{p}_{4,e}={\begin{bmatrix}\text{b} & -\text{b}&0 \end{bmatrix}}^T$.

Then, the velocity $^b\bm{v}_{i,e}$ is converted to the $X_{wi}O_{wi}Y_{wi}$ frame.
For CuRobot, the motion in the $O_{wi}X_{wi}$ direction cannot be provided by the rotation of 
the wheel. Therefore, only its component in the $O_{wi}Y_{wi}$ direction is calculated as follows:

\begin{equation}
        ^{wi}v^y_{i,e} = ^bv^x_{i,e} cos\upphi_i + ^bv^y_{i,e} sin\upphi_i ,
\end{equation}
where $\upphi_i=\upgamma_i + \frac{\pi}{2}$ is the angle between $O_{wi}Y_{wi}$ and $O_bX_b$.

Finally, the rotational speed of each wheel can be formulated as follows:

\begin{equation}
        \dot{q}_i = \frac{^{wi}v^t_{i,e}}{r_e}.
\end{equation}

By combining the equations above and writing them in matrix form, the speed of the four driving wheels can be expressed as:
\begin{equation}
        \bm{\dot{q}} = diag\left\{ \mathbf{B}(\bm{v}_m + \bm{\varOmega}\mathbf{P}) \right\} / r_e ,
\end{equation}
where $\bm{\dot{q}} =\left[\dot{q}_1 \quad \dot{q}_2 \quad\dot{q}_3 \quad\dot{q}_4 \right]$ is the rotational speed of the four driving wheels, 
$ \bm{\Omega} = \left[\bm{\omega}_m\right] = \begin{bmatrix}
        0 & -\omega_m & 0 \\ \omega_m & 0 & 0 \\ 0 & 0 & 0 
\end{bmatrix}$ is the antisymmetric matrix of the vector $\bm{\omega}_m = {\left[0 \quad 0 \quad \omega_m \right]}^T$,
       $ \mathbf{B} = \begin{bmatrix}
        cos\upphi_1 & sin\upphi_1 & 0 \\ 
        cos\upphi_2 & sin\upphi_2 & 0 \\
        cos\upphi_3 & sin\upphi_3 & 0 \\
        cos\upphi_4 & sin\upphi_4 & 0 
\end{bmatrix}$ is used to solve the component of the wheel velocity on $O_{wi}X_{wi}$, and 
$\mathbf{P} =\left[\mathbf{p}_1 \quad \mathbf{p}_2 \quad \mathbf{p}_3 \quad \mathbf{p}_4 \right]$.

\subsection{Attitude Redefinition}

We redefine the direction of the body frame when the robot flips over or when the tilt angle is too large. For clarity, we will refer to the body frame defined above as the initial frame, and the six faces of the cube are marked with the numbers 1-6, as shown in Fig. \ref{fig:face}. We use an accelerometer to determine which face is the landing face. The gravitational acceleration is decomposed onto the three axes of the body frame, and the magnitudes of the absolute values of the three components are compared. The two surfaces perpendicular to the axis of the component with the largest absolute value are the possible landing faces, and the landing face can then be determined by the plus or minus sign of that component. Define the direction in which the robot's CoM points toward the ground as the $O_bZ_b$ direction of the new body frame. 
When the landing face is 2, 3, or 4, the new $O_bX_b$ is the same as the $O_bX_b$ of the initial frame; When the landing face is 5, the new $O_bX_b$ direction is the same as the $-O_bZ_b$ direction of the initial frame; And when facing the ground is 6, the new $O_bX_b$ is the same as $-O_bZ_b$ direction of the initial frame. The new $O_bY_b$ direction of initial frame is determined according to the right-hand rule, and in this way, we have defined the body frame for all the faces as landing faces.

When the body frame changes, we need to compute the new body frame's orientation with respect to the world frame, since the pose acquired by the sensors, denoted as attitude matrix $\bm{R}_{3\times3}$ is always the pose of the initial body frame about the world frame. We denote the elements of $\bm{R}_{3\times3}$ by $r_{ij}$ (where $i$ and $j$ range from 1 to 3). According to the definition, the yaw angle can be calculated by the following equation:

\begin{equation}
        \psi = \arctan (r_{12}/r_{11})
\end{equation}
Therefore, for the redefined body frame, the yaw angles for different landing faces are:

\begin{equation}
        \psi = \begin{cases} \arctan (r_{21}/r_{11}), \quad & face = 1,2,3 \quad and \quad 4 \\
                             \arctan (-r_{23}/-r_{13}),& face = 5 \\
                             \arctan (r_{23}/r_{13}), &face = 6 \end{cases}
\end{equation}

\subsection{Controller Design}

\begin{figure}[ht]
    \centering
    \includegraphics[scale=0.35]{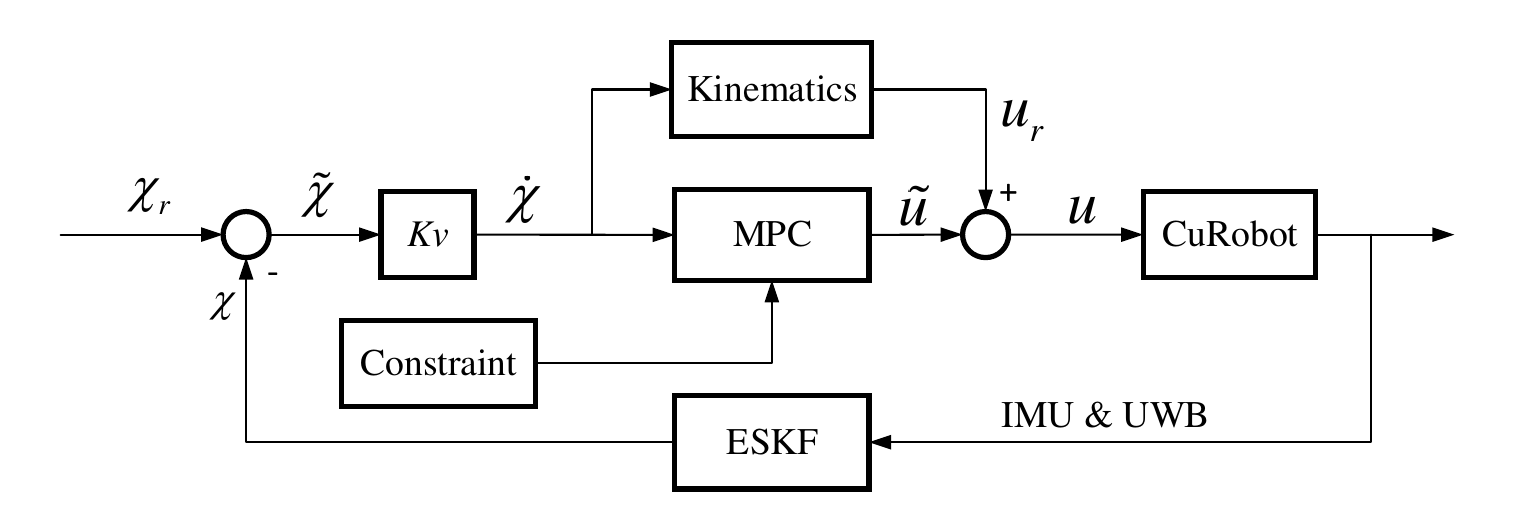}
    \caption{The controller framework}
    \label{fig:control_system}
\end{figure}

First, we represent the desired trajectory as a time series of states $\bm{\chi}_r(t)$, and then calculate the motion speed $\dot{\bm{\chi}}$ of the robot from the state deviation $\widetilde{\bm{\chi}}$ of the desired state from the actual state. Next, the reference speed $\bm{u}_r$ of the drive wheel is calculated based on the kinematic model. Finally, we add the correction $\widetilde{\bm{u}}$ calculated by the MPC to get the final value of the wheel speed $\bm{u}$.
In this paper, we want to avoid large differences in wheel speed, so we design the following objective function:

\begin{equation}
        \label{objective_function}
        \begin{split}
        J(k)=\bm{X}_k^T \overline{\bm{Q}} \bm{X}_k + \bm{U}_k^T \overline{\bm{R}} \bm{U}_k + \bm{U}_k^T \overline{\bm{E}} \bm{U}_k + 2\bm{U}^T_r \overline{\bm{E}} \bm{U}_k + \bm{U}^T_r \overline{\bm{E}} \bm{U}_r
\end{split}
\end{equation}
The last three terms contain the difference between the speeds of each driving wheel. This objective function minimises the difference in the rotational speed of each driving wheel.
We estimate the position of the robot as feedback information in the entire control system by fusing the 
UWB and IMU data with the error-state Kalman Filter (ESKF) algorithm \cite{ref30}.
The controller's framework is shown in Fig. \ref{fig:control_system}.

\section{RESULTS AND DISCUSSION}
In order to verify the ability of the robot to continue to move after tipping over and the control effect 
of the trajectory tracking controller, simulation experiments and physical experiments were conducted. The 
simulation experiments were implemented in the webots environment.

The parameters of the simulated environment, including the robot's mass and the friction coefficient between the wheel and the ground, are presented in Table \ref{map-coefficent}. The steering direction of each wheel was predefined as counterclockwise and considered positive when viewed from the outside towards the center of the cube.

\begin{table}[!htb]
        \centering
        \caption{Parameters of the simulation experiments}
        \label{map-coefficent}
        \begin{tabular}{cc}
        \hline
        param  & value  \\ \hline
        mass &  3.0 kg   \\
        $r_e$ & 0.05 m  \\
        b  &  0.01 m \\
        friction coefficient & 500 \\
        contact stiffness & 10 N/mm \\
        contact damping & 5 N/(mm/s)\\
          \hline
        \end{tabular}
\end{table}

To prevent significant deviation from the reference trajectory, the robot's position was collected as 
feedback during the simulation. A proportional-integral-derivative controller was used to calculate the 
desired velocity of the robot, and the motor speed was determined based on the kinematic model.

\subsection{Tumble Motion Experiment} 
In the tumbling motion experiment, we make the robot tumble with the help of a slope. During the experiment, the robot uses an attitude sensor to determine which side is in contact with the ground and then sets the four wheels on that side as the driving wheels. 

\begin{figure}
       \centering
        \includegraphics[scale=0.6]{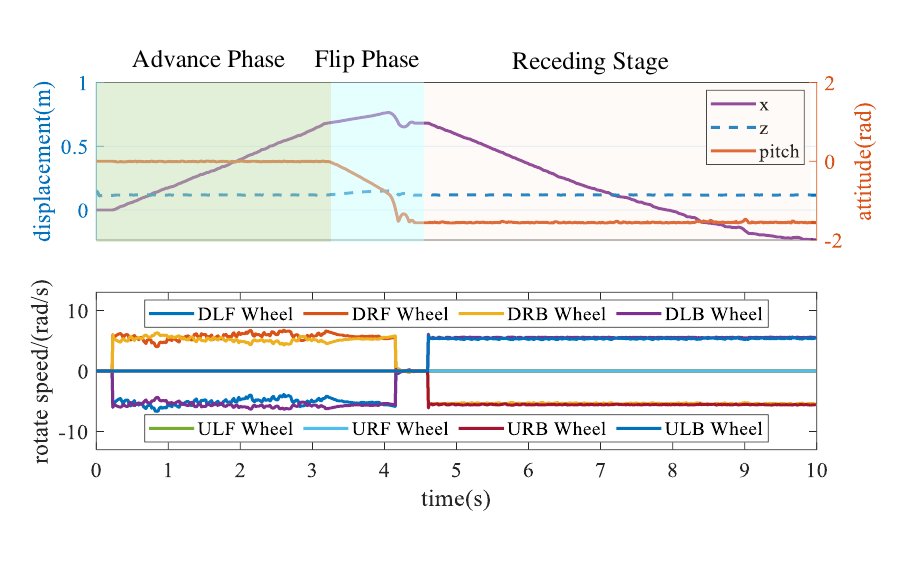}
        \caption{Simulation result of tumble motion experiment}
        \label{flip_sim}
\end{figure}

The results of the experiment are depicted in Fig. \ref{flip_sim}, where the blue zone represents the flipping process. Initially, the robot initiated its journey from a level surface and encountered a slope at t = 3.2 s. As the robot's front wheel transitioned onto the slope, the robot's body began to tilt. Subsequently, due to the steepness of the slope, the robot tipped over, eventually coming to rest on the adjacent side. Upon reorienting itself to establish contact with the ground, the robot moved away from the slope in the opposite direction. Throughout this phase, the pitch angle underwent a transition from 0° to -90°, and the robot's height exhibited variations, initially showing a gradual increase and later rapidly descending to its original level.

The bottom of Fig. \ref{flip_sim} shows the speed of the drive wheels, and it can be seen that the drive wheels are different before and after the robot is flipped, i.e., the drive wheels are DLF, DRF, DRB, and DLB wheels before flipping, and DRB, DLB, URB, and ULB wheels after flipping.

\subsection{Line Trajectories Tracking}

We designed the linear trajectory tracking experiment to analyze the relationship between the robot's motion direction and the accuracy of tracking predefined trajectory. In light of CuRobot's symmetrical structure, the reference trajectory's orientation relative to the ${OX}$ axis, denoted as $\theta$, commences at 0° and is incrementally adjusted in 5° intervals until it reaches 45°.
In addition, all the trajectories have a length of 8 meters and the robot is moving at a speed of 0.1m/s.  

\begin{figure*}[!t]
        \centering
        \subfloat[experiment schematic]{\includegraphics[scale=0.35]{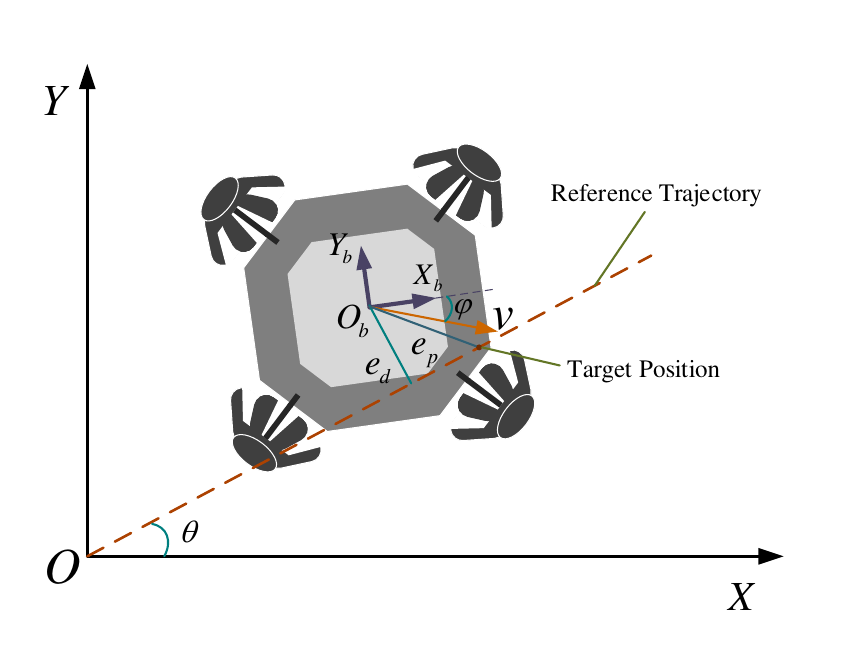}
        \label{fig:traj_define}}
        \subfloat[experiment trajectories]{\includegraphics[scale=0.28]{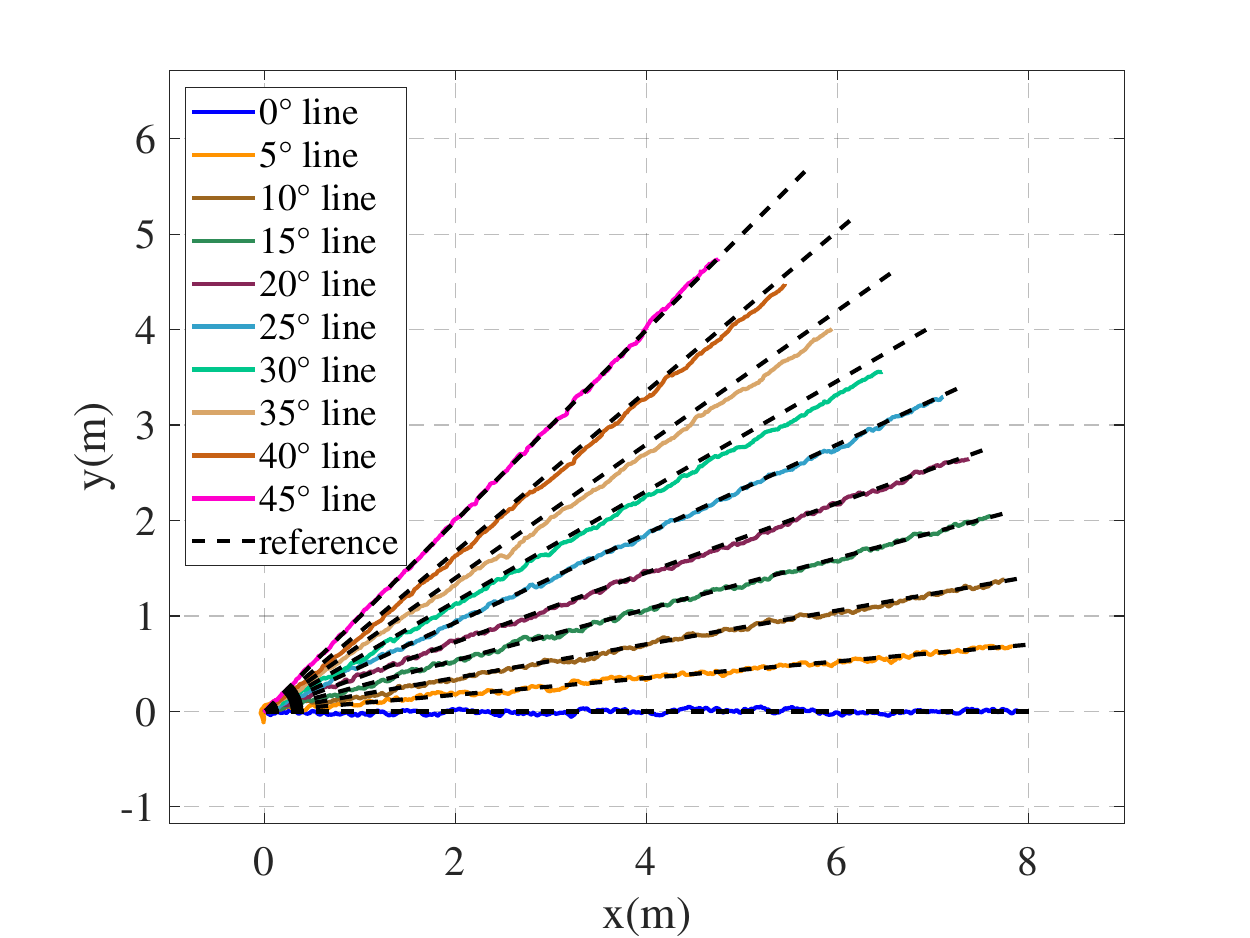}
        \label{trajectory}}
        \hfil
        \subfloat[results statistics]{\includegraphics[scale=0.28]{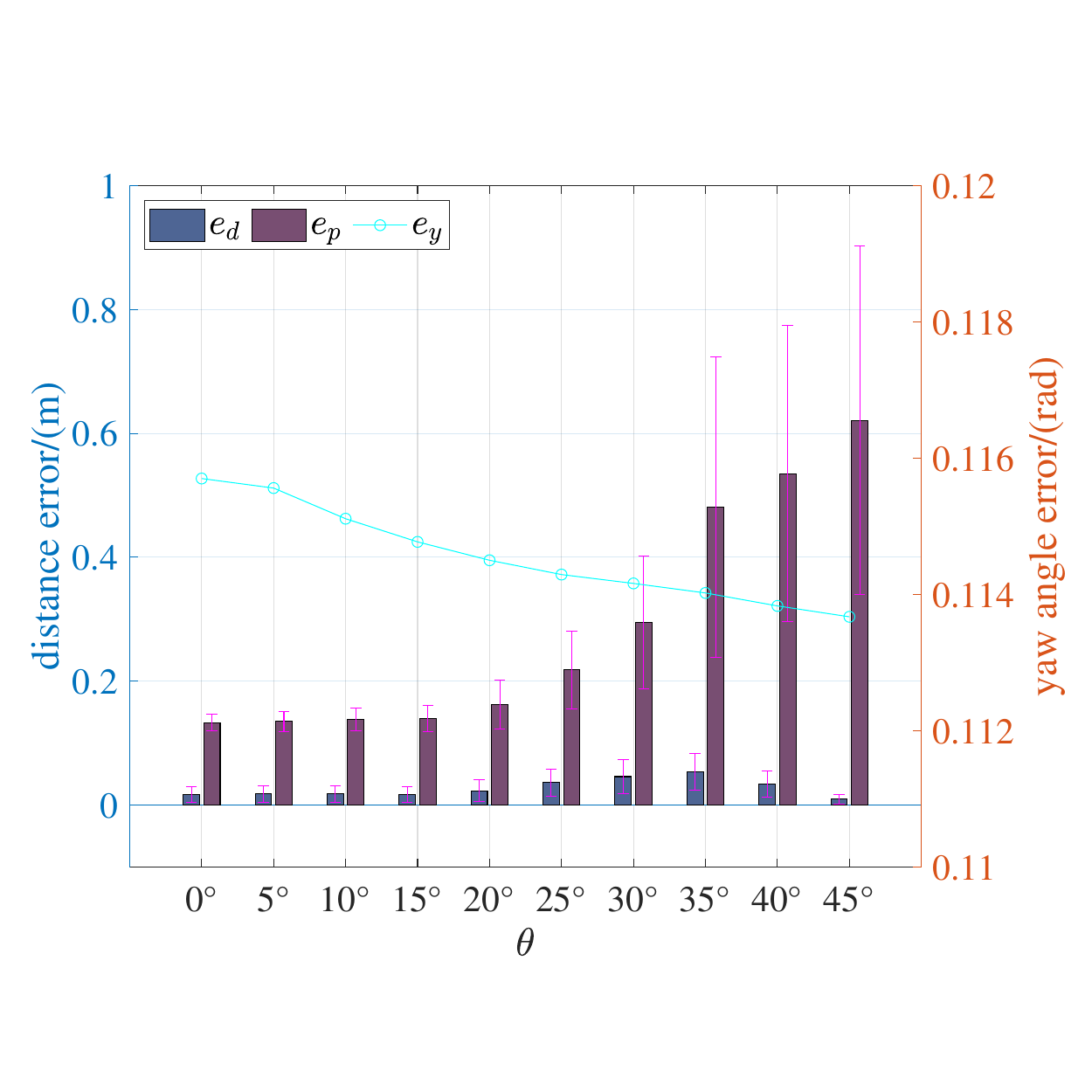}
        \label{bar}}
        \label{line_trajectory}
        \caption{line trajectory}
\end{figure*}

The robot's initial position aligns with the starting point of the reference trajectory, with a constant yaw angle of $\psi$ maintained at 0° throughout its motion. Consequently, the angle $\varphi$, which represents the robot's velocity relative to ${O_bX_b}$, is identical to $\theta$.
The experiments tracking each Trajectory were performed 10 times repetitively, and we counted three 
metrics: position error $e_p$, off-trajectory distance error $e_d$, and yaw angle error $e_y$, as illustrated in Fig. \ref{fig:traj_define}.

From Fig. \ref{bar}, it can be seen that when the angle $\theta$ is below 20°, the position error $e_p$ remains nearly constant; however, as $\theta$ surpasses 20°, the position error $e_p$ progressively escalates with increasing values of $\theta$.
The distance error $e_d$ exhibits an upward trend until $\theta$ reaches 35°, after which it begins to decrease. As $\theta$ reaches 45°, $e_d$ converges to nearly zero. This result can be attributed to the fact that, at this point, the two wheels on one diagonal have ceased their rotational movement, functioning solely to provide friction in the direction opposing the robot's motion. Simultaneously, the two wheels on the other diagonal maintain identical rotational speeds, thereby minimizing relative sliding between the wheels and the ground.
As for the yaw angle error $e_y$, it also decreases as the $\theta$ angle increases, but its absolute value is so small that it can be ignored.

\subsection{Planar Omnidirectional Motion Experiment}
We set up three circular trajectories tracking experiments, and in all experiments, the robot was asked to track the same circle trajectories while with different trajectories of $\psi$, namely

\begin{equation}
        \begin{cases} \psi =0,\\ \psi = \arctan \frac{\dot s_y}{\dot s_x},  \\ \dot \psi = 0.1 \quad rad/s. \end{cases}
 \nonumber
\end{equation}
The duration of completing one lap of the circular trajectory is $20\pi(\approx  62.8)$ seconds, and we set the time for each simulation experiment to 100 seconds. 

\begin{figure*}[!t]
        \centering
        \label{circle_result}
        \subfloat[trajectory]{\includegraphics[width=2.3in]{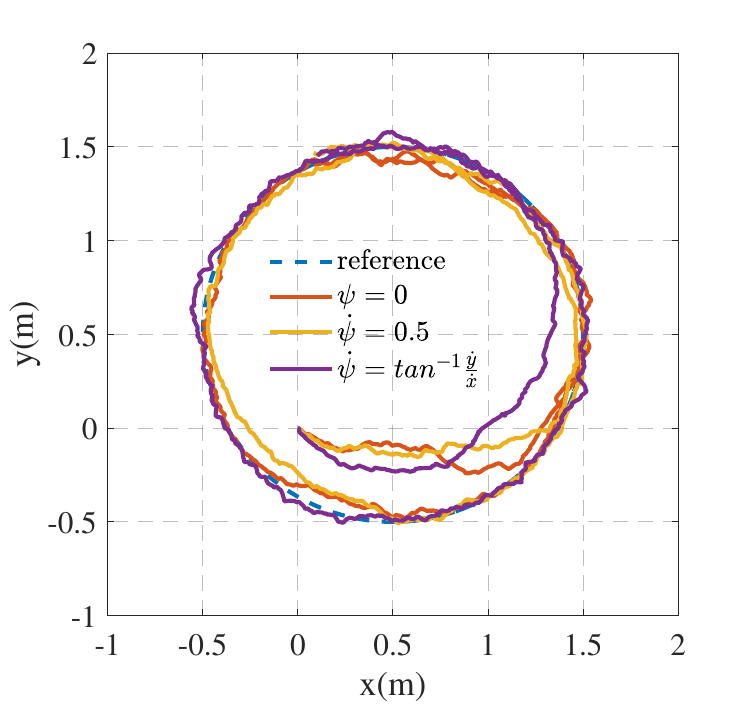}%
        \label{fig:s_circle_traj_f}}
        \hfil
        \subfloat[state]{\includegraphics[width=2.3in]{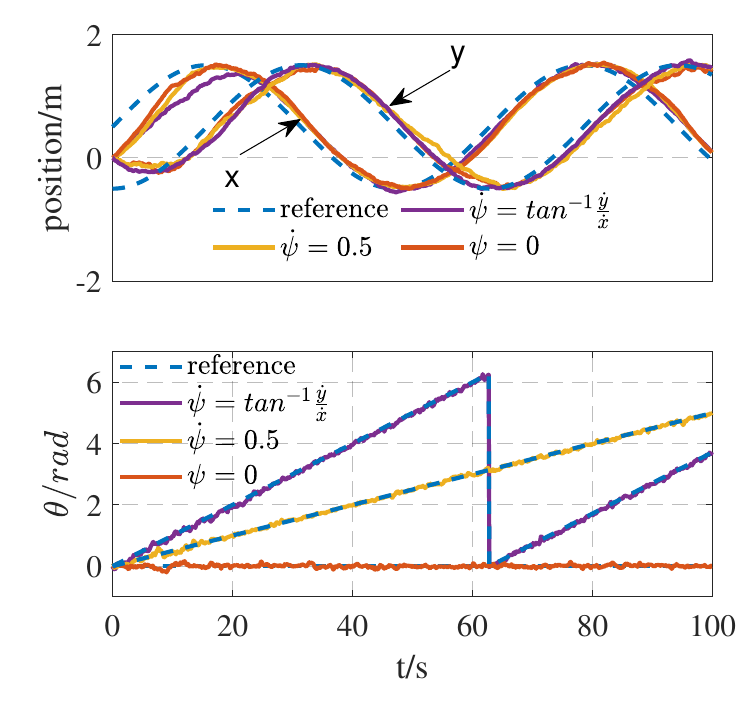}%
        \label{fig:s_circle_state_f}}
        \hfil
        \subfloat[wheel speed]{\includegraphics[width=2.3in]{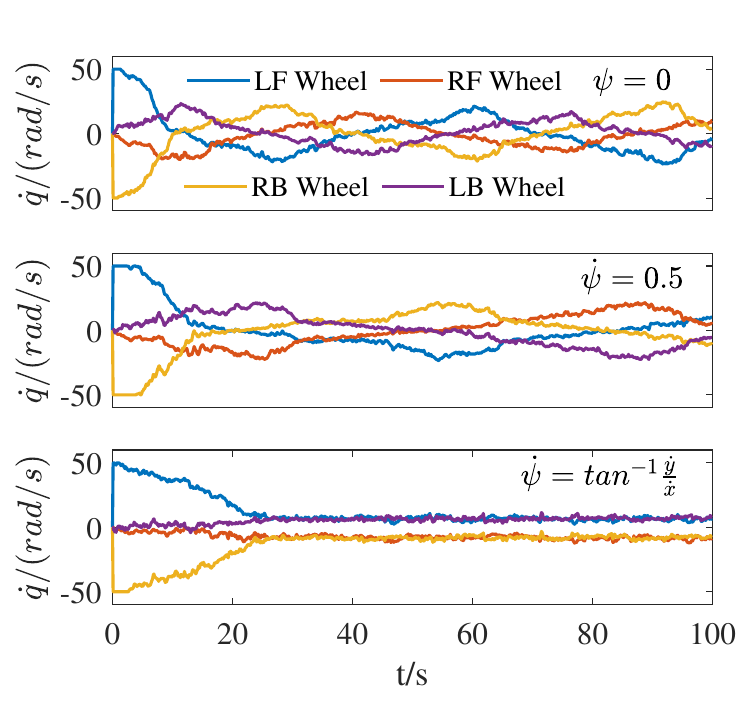}
        \label{fig:circle_f_sim_u}}
        \caption{Results of tracking circular trajectory under different yaw angle settings.}
\end{figure*}

The reference trajectory and actual robot trajectory are shown in Fig. \ref{fig:s_circle_traj_f}.
The initial position of the robot does not coincide with the reference trajectory; however, after a certain duration, the robot successfully catches up with the reference trajectory (defined in this document as the distance between the actual position and the reference position is less than 0.2m). Notably, the time required for the robot to achieve this alignment when its yaw angle matches the direction of motion is approximately 20 seconds, which exceeds the time needed for the other two experiments, which is less than 10 seconds, but once tracked on the reference trajectory, it deviates minimally from a reference trajectory.

It can be seen from Fig. \ref{fig:circle_f_sim_u} that the speed of the driving wheel has symmetry, that is, for circular motion, the amplitude of the speed of the two driving wheels on the diagonal is equal, which is consistent with the kinematic model. At the beginning of the experiment, the drive wheels reach their maximum value to ensure rapid catch up with the reference trajectory from the initial location.

\subsection{Trajectory Tracking on Uneven Terrain}
In this experiment, the robot followed an eight-shaped trajectory, with the yaw angle aligned with its motion direction, traversing an uneven terrain. The terrain is a circular pit with a diameter of 3 meters and the terrain surface was made rugged using Berlin noise. 

\begin{figure*}[!t]
        \centering
        \subfloat[trajectory]{\includegraphics[width=2.3in]{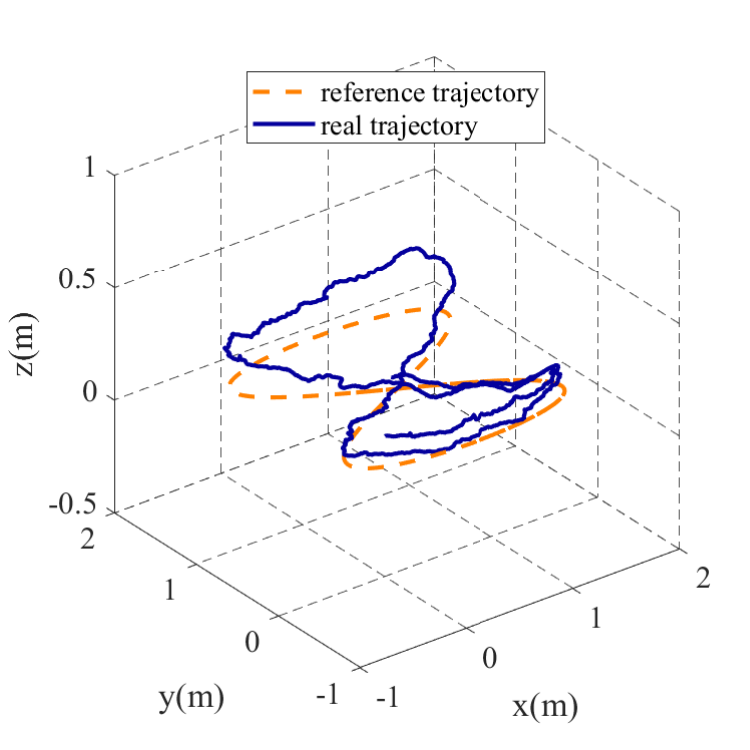}
        \label{fig:s_eight_traj_v}}
        \subfloat[state]{\includegraphics[width=2.3in]{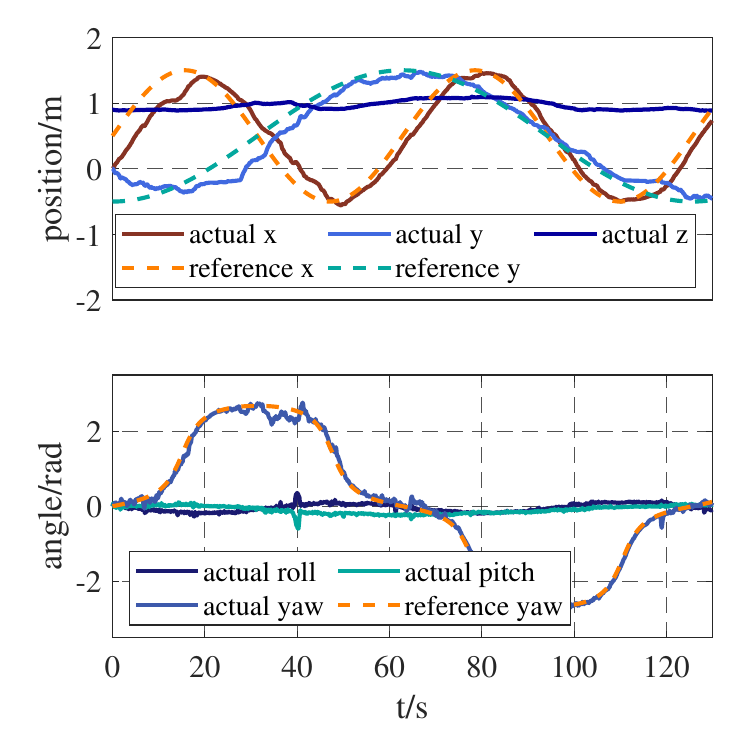}
        \label{fig:s_eight_state_v}}
        \subfloat[wheel speed]{\includegraphics[width=2.3in]{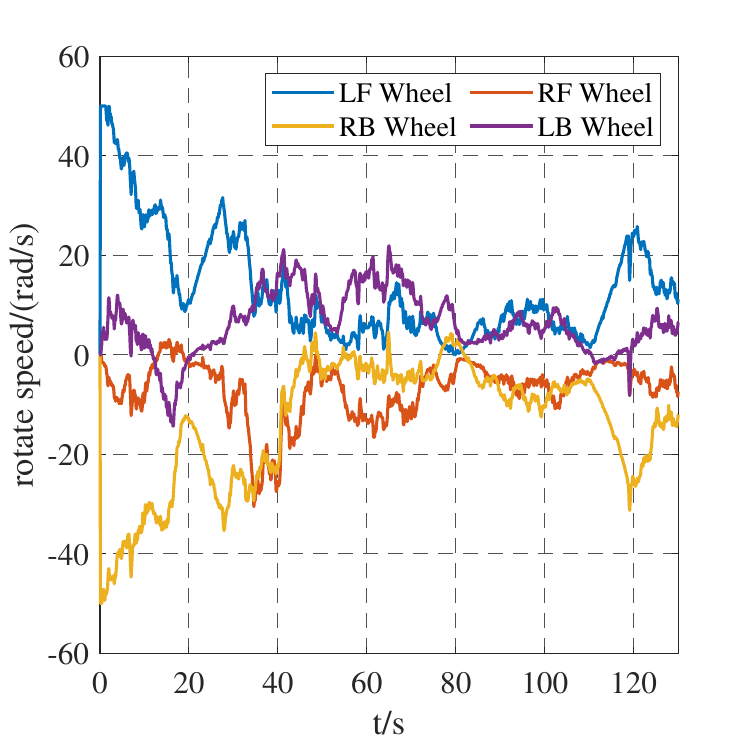}
        \label{fig:s_eight_u_v}}
        \caption{Results of tracking eight-shaped trajectory on uneven terrain.}
        \label{eight_result}
\end{figure*}

As evident from Fig. \ref{fig:s_eight_state_v}, throughout the entire motion, there are persistent fluctuations in the height of the robot's center of mass, as well as in its roll and pitch angles, attributable to the uneven terrain.

The maximum difference in the robot center of mass height is 0.28 m and the maximum tilt angle (pitch or roll) of the 
robot is 0.59 rad. The robot is able to track the reference trajectory in general, but the tracking accuracy is not as 
great as on a flat surface. Fig. \ref{fig:s_eight_u_v} illustrates the rotational speed of the wheels, which are 
symmetric but fluctuate more dramatically than when moving on a flat surface.

\subsection{Trajectory Tracking With Flipping} 
\begin{figure*}[!t]
        \centering
        \subfloat[trajectory]{\includegraphics[width=2.3in]{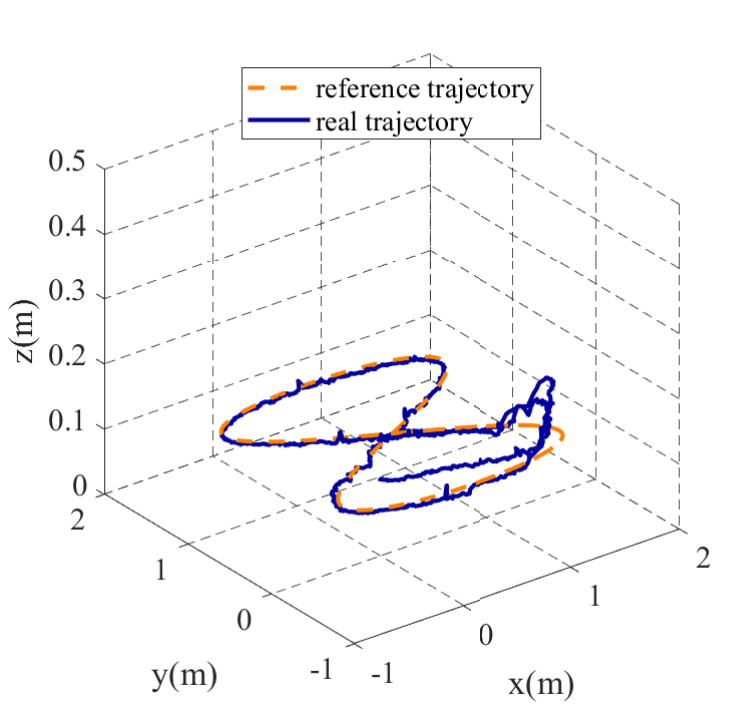}
        \label{fig:s_eight_un_traj_v}}
        \hfil
        \subfloat[state]{\includegraphics[width=2.3in]{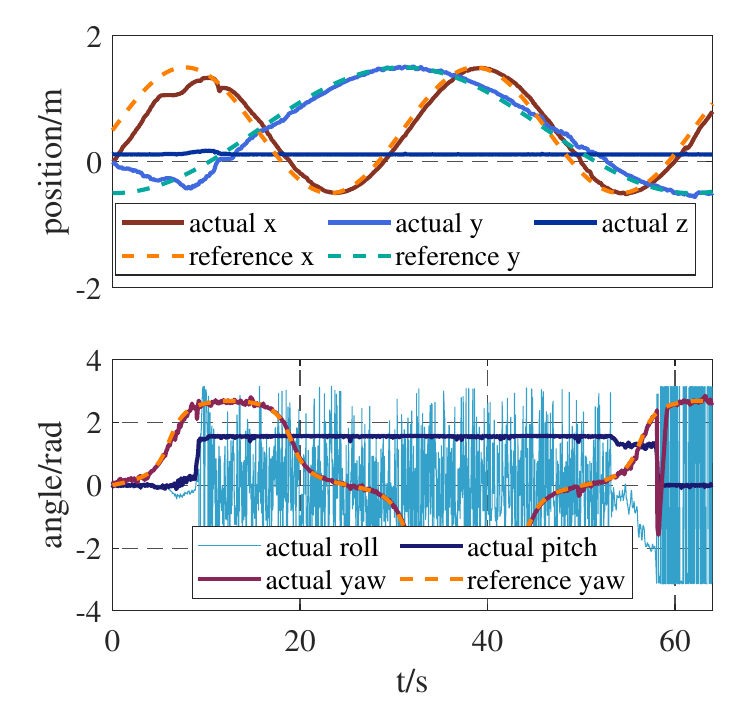}
        \label{fig:s_eight_un_state_v}}
        \hfil
        \subfloat[wheel speed]{\includegraphics[width=2.3in]{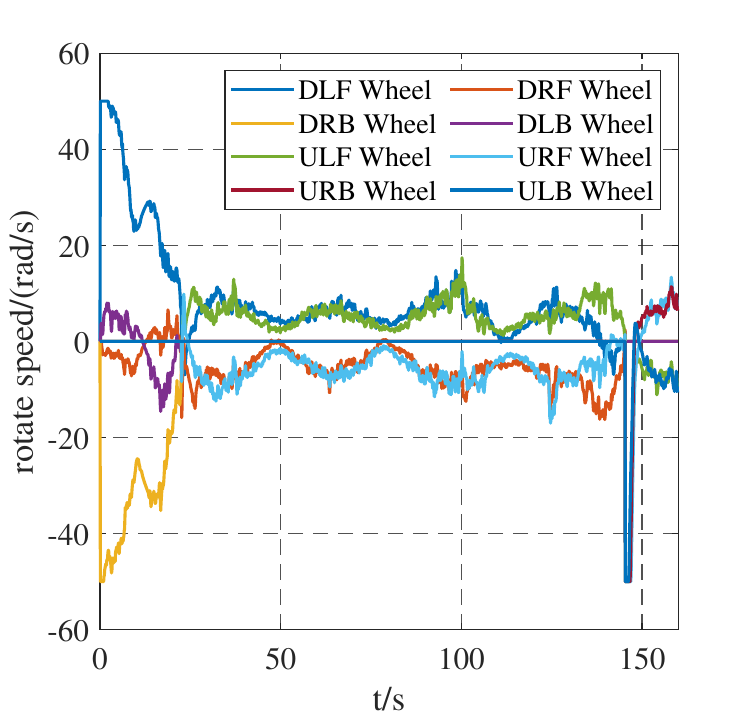}
        \label{fig:eight_un_v_sim_u}}
        \caption{Results of tracking eight-shaped trajectory during which the robot flips over.}
        \label{eight_un_result}
\end{figure*}

In this experiment, we placed a 15° slope on the flat ground and then similarly had the robot track an 
eight-shaped trajectory. The robot first climbs up the ramp, then falls down from the side, and changes its landing face. As shown in Fig. \ref{fig:s_eight_un_state_v}, it can be seen from the change in the pitch angle of the robot that the robot flipped a total of two times at t=10s and t=115s, and the landing face changed to 6 and 3, respectively.

In the first stage of motion, the driving wheels are the four wheels on face 1, i.e., 
DLF, DRF, DRB, and DLB wheels. When the robot flips over and lands on face 6, the driving wheels become DLF, DRF, ULF, and URF wheels, and after the second flip, the driving wheels become ULF, URF, URB, and ULB wheels.
After the second flip, there was a sharp change in yaw angle, and under the action of the controller, it quickly returned to the desired value, which also led to the speed of the driving wheels becoming saturated, as shown in Fig. \ref{fig:s_eight_un_state_v} and Fig. \ref{fig:eight_un_v_sim_u}.

\subsection{Prototype Experiment}
\begin{figure*}
        \centering
        \subfloat[flipping motion]{\includegraphics[scale=0.3]{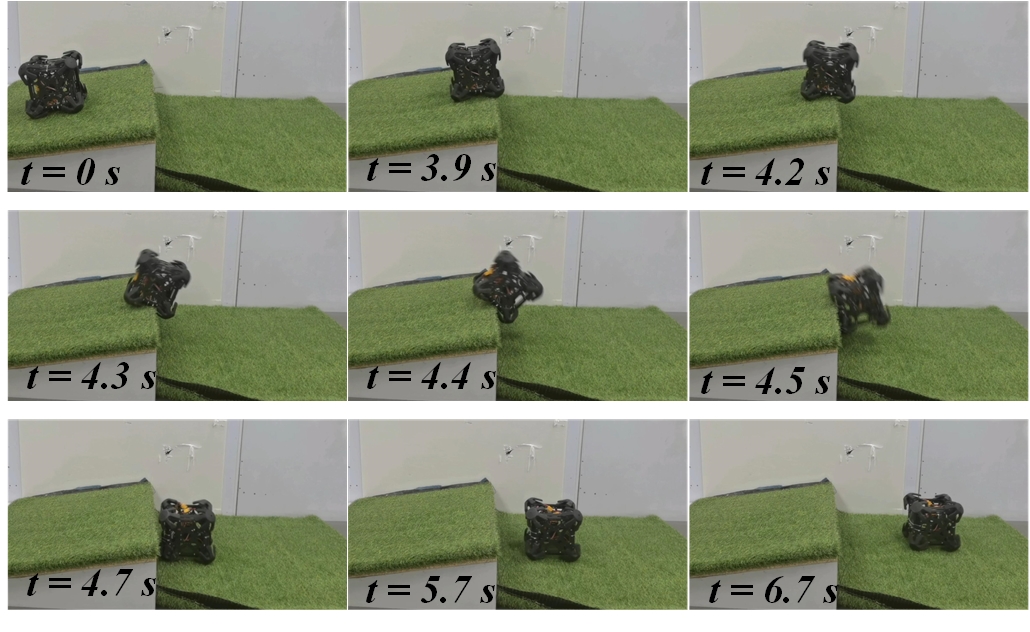}
        \label{fig:flip_prototype}}
        \subfloat[tracking circle trajectory]{\includegraphics[scale=0.3]{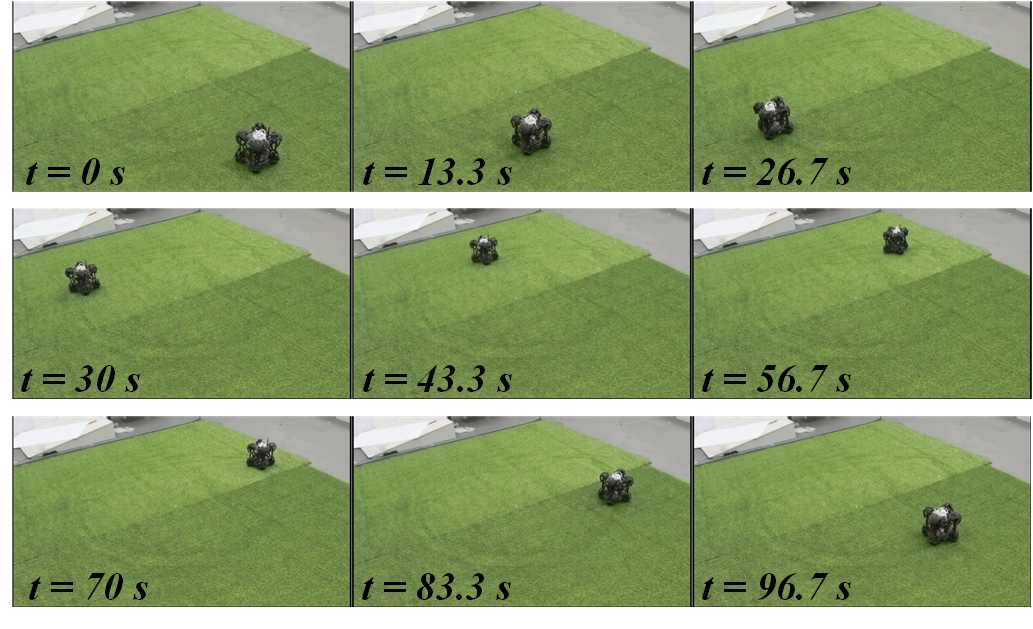}
        \label{fig:circle_process_prototype}}
       \caption{Prototype experiment.}
       \label{fig:prototype}
\end{figure*}

In this section, we performed two prototype experiments, the first of which was to have the robot travelling towards the edge of a slope and then fall, and the fall lasting about 0.5 seconds (4.2s-4.7s). The robot's landing face then changed and reset the drive wheels to continue travelling.
The second experiment involved having the robot track a circular trajectory of 2 meters in diameter at a fixed yaw angle clockwise. The experimental procedure lasted about 100 seconds, during which the robot deviated from the intended trajectory, but was able to quickly return to the intended trajectory with the help of the controller.
The screenshot of the experimental procedure is shown in Fig. \ref{fig:prototype}.

\section{CONCLUSIONS}
In this work, we firstly propose a novel reversible robot with a cubic structure and driven by eight conical 
wheels. Further, we establish its kinematic model and present a redefinition of the post-flip attitude. Later, 
we design a trajectory tracking controller based on MPC. Afterword, our simulation experiments comprehensively examine 
the correlation between trajectory tracking accuracy and the direction of motion. Finally, it is validated that the 
designed controller enables the robot to accurately track a predetermined trajectory and that the robot can continue 
to traverse and regain its previous trajectory even after it flips over.




\end{document}